\documentclass[final]{cvpr}

\usepackage{times}
\usepackage{epsfig}
\usepackage{graphicx}
\usepackage{amsmath}
\usepackage{amssymb}
\usepackage{multirow}
\usepackage{subfig}

\usepackage[pagebackref=true,breaklinks=true,colorlinks,bookmarks=false]{hyperref}

\def\cvprPaperID{3} % *** Enter the CVPR Paper ID here
\def\confYear{CVPR 2021}

\begin{document}

%%%%%%%%% TITLE
\title{SyDog: A Synthetic Dog Dataset for Improved 2D Pose Estimation}

\author{Moira Shooter \qquad Charles Malleson \qquad Adrian Hilton \\
University of Surrey\\
Stag Hill, University Campus, Guildford GU2 7XH\\
{\tt\small \{m.shooter,charles.malleson,a.hilton\}@surrey.ac.uk}
}

\maketitle

%%%%%%%%% ABSTRACT
\begin{abstract}
Estimating the pose of animals can facilitate the understanding of animal motion which is fundamental in disciplines such as biomechanics, neuroscience, ethology, robotics and the entertainment industry. Human pose estimation models have achieved high performance due to the huge amount of training data available. Achieving the same results for animal pose estimation is challenging due to the lack of animal pose datasets. To address this problem we introduce SyDog: a synthetic dataset of dogs containing ground truth pose and bounding box coordinates which was generated using the game engine, Unity.
We demonstrate that pose estimation models trained on SyDog achieve better performance than models trained purely on real data and significantly reduce the need for the labour intensive labelling of images. We release the SyDog dataset as a training and evaluation benchmark for research in animal motion. 
\end{abstract}
 %In order to show the benefit of using the synthetic data generated by our system, we estimate the two dimensional pose of dogs in images. %Deep learning models trained purely on synthetic data often tend to fail to generalize to real world cases. In order to bridge the domain gap we run two experiments; one where we apply transfer learning and the other where we use a mixed dataset (real+synthetic). 

%-------------------------------------------------------------------------
%%%%%%%%% BODY TEXT
%-------------------------------------------------------------------------
\section{Introduction}
\begin{figure*}
\begin{center}
\includegraphics[width=0.8\linewidth]{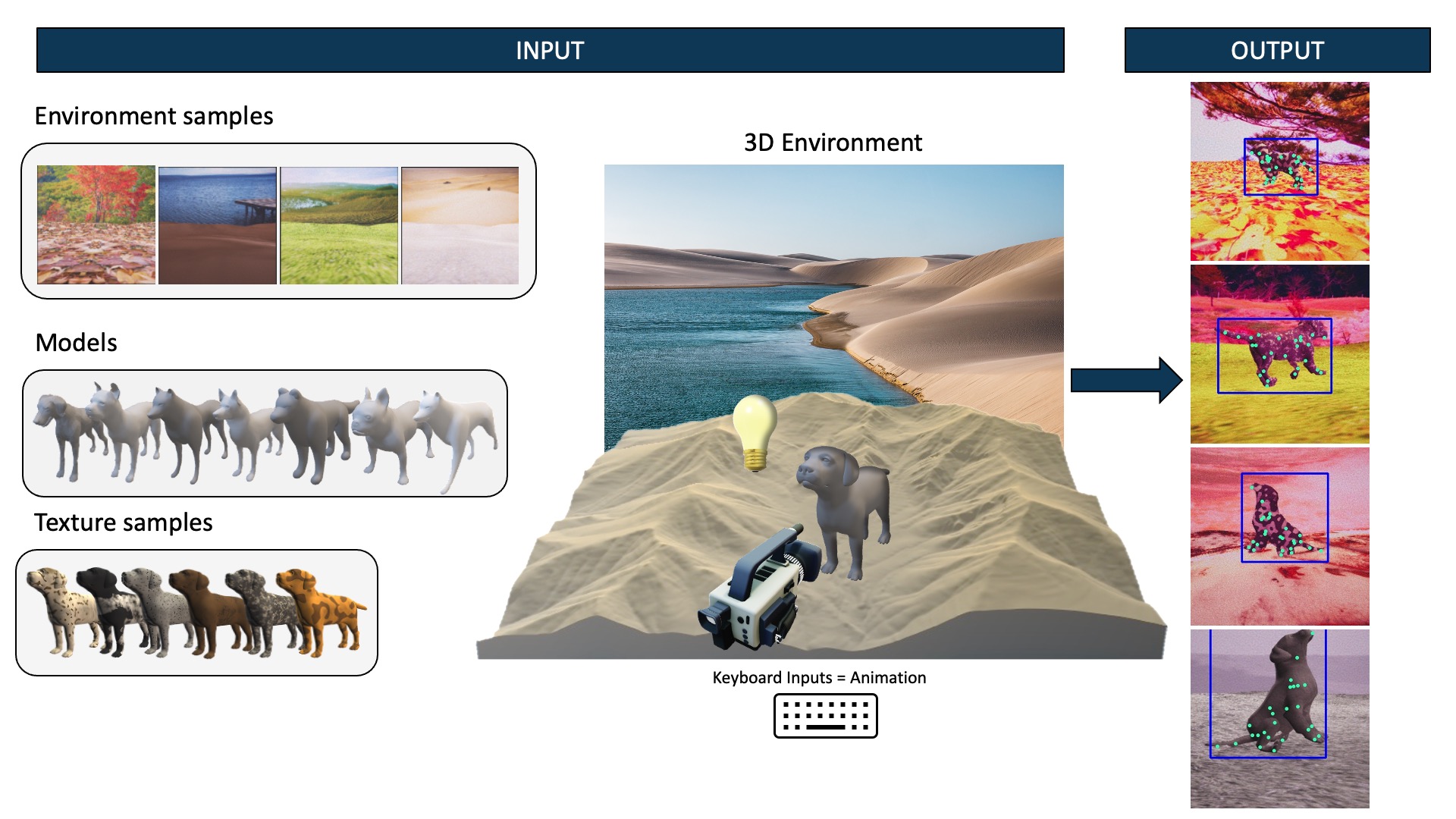}
\end{center}
   \caption{Pipeline showing the process of generating the SyDog dataset. The dog's motion is controlled using keyboard inputs. A virtual camera follows and renders a frame of the dog with different appearance, pose, lighting (post-processing effects), environment, and camera view points. These parameters are randomly sampled to make the data more diverse. RGB images, 2D pose and bounding box coordinate annotation are simultaneously generated.}
\label{fig:pipeline_overview}
\end{figure*}
%Pose estimation is a computer vision problem where information about skeleton configuration is deduced by locating joint positions in visual data such as images and videos. 
Estimating the pose of animals from video \cite{Gosztolai2020.09.18.292680, graving2019deepposekit, Mathis2018, Pereira331181} helps to understand the animal motion and this supports many applications and disciplines such as veterinary science where lameness can be diagnosed early and recovery monitored; biomechanical applications where gait is analysed to improve animal performance in sports such as horse racing and dressage; neuroscience where motion is analysed to understand behaviour and/or relate motion to brain activity \cite{info:doi/10.2196/16194}; robotics where robots learn from animal motion data \cite{RoboImitationPeng20}; and in the entertainment industry to produce more natural and realistic animal animations and to create 3D representations of animals. The traditional and most accurate method to track the motion of subjects of interest is optical motion capture. This involves placing reflective markers on the subject and uses a system with multiple cameras to capture their 3D location. This method has its disadvantages in that it requires expertise and time to set up, it can be stressful to the animal, it can change the animal's behaviour, animals can be uncooperative and in some cases it is impossible to bring the animal into a lab. Another disadvantage is that the lighting conditions need to be fairly controlled, typically restricting such systems to laboratories. Non-contact video-based estimation of animal motion has the potential to overcome these limitations. Deep learning methods are known to perform well with huge amounts of data. The main focus in the literature has been on human pose estimation \cite{DBLP:journals/corr/abs-1812-08008, DBLP:journals/corr/FangXL16, DBLP:journals/corr/HeGDG17,
DBLP:journals/corr/NewellYD16,DBLP:journals/corr/abs-1902-09212, DBLP:journals/corr/TompsonGJLB14, DBLP:journals/corr/ToshevS13} where large amounts of training data have allowed high accuracy to be obtained. It is challenging to achieve the same quality results for animals as there is less training data available \cite{biggs2020left, biggs2018creatures,DBLP:journals/corr/abs-1908-05806,Kearney_2020_CVPR}. The standard way to create datasets is to annotate each image manually, but annotating several keypoints in thousands of images is both labour intensive and expensive. However, in recent years the generation of synthetic data has been an accelerator for machine learning \cite{DBLP:journals/corr/ChenWLSTLCC16,DBLP:journals/corr/abs-1912-08265,DBLP:journals/corr/abs-1910-07113,DBLP:journals/corr/RichterVRK16,DBLP:journals/corr/abs-1908-07201}. %In computer vision, synthetic data can be generated in many ways such as by rendering photorealistic objects onto different backgrounds or rendering the objects including the scene from a virtual scene using 3D software.

In this work we address the lack of animal datasets by creating a dataset consisting of images of dogs rendered using a real-time game engine. 
% In this work we evaluate the use of our synthetic dataset by estimating the two-dimensional (2D) pose of dogs in images. To address the lack of animal datasets, we create synthetic data containing dogs in real-time to train pose estimation models. 
%The system was built upon Zhang \etal project \cite{10.1145/3197517.3201366}. By using this system we were able to create a variety of dog poses. 
To add variation into the data, the dog's appearance and pose, the environment, the camera viewing points and the lighting conditions were modified. Using this approach, we generated a synthetic dataset containing 32k annotated images.% which took 33 milliseconds per frame. 
We evaluate the pose estimation models trained with synthetic data on the StanfordExtra dataset \cite{biggs2020left}. Because networks trained only on synthetic data often fail to generalize to real world examples (the domain gap) \cite{DBLP:journals/corr/BarbosaCCRT17,DBLP:journals/corr/abs-1807-01990}, two techniques were applied separately: fine-tuning the networks and training the networks with a combination of real and synthetic samples. %We conducted six experiments to evaluate the effectiveness of models trained on our synthetic dataset.
We demonstrate that models trained on synthetic data increase the models' performances and reduce the need for labour intensive image annotation. 

The \textbf{main contributions} of this work are: (i) We present a real-time system that generates 2D annotated images containing dogs. (ii) We release SyDog, a large scale annotated dataset of dogs with 2D keypoints and bounding box coordinates. (iii) We show that using the SyDog dataset improves the accuracy of pose estimation models and reduces the need for labour intensive labelling.
%The SyDog is made available at: \url{https://github.com/mshooter/SyntheticDog}
% include related work (old version)
%-------------------------------------------------------------------------
%%%% DATA GENERATION %%%%%
%-------------------------------------------------------------------------
\section{Data Generation}
In this section we present how the SyDog dataset was generated. Figure \ref{fig:pipeline_overview} demonstrates the pipeline overview for generating the synthetic data.

We generated \textbf{the synthetic dataset} using the game engine Unity3D. We built upon Zhang \etal's project \cite{10.1145/3197517.3201366} which produces natural animations for quadruped animals from real motion data using a novel neural network architecture which they call Adaptive Neural Networks. By using this system we were able to control the animal's motion using keyboard inputs and make the dataset more varied by transitioning the dog's pose from one state to another. To add more variety into the dataset, the dog’s appearance, the environment, the camera viewing points and the lighting conditions (post-processing effects) were randomly modified. We produced 32k images along with annotations of 25 keypoints and bounding box coordinates. We refer the reader to the supplementary material for samples of the SyDog dataset.
%------------------------
\\\textbf{Dog models.} We used 8 different type of dogs, 1 came with Zhang \etal's project, which we will refer to as the \textit{default} model, 5 were imported from the RGBD-Dog dataset \cite{Kearney_2020_CVPR}, and 2 were a fat and a skinny version of the \textit{default} model. The models represent dogs ranging from big to small sized breeds. The models were manually scaled and rigged based on the \textit{default} model for the models to be correctly imported into the project.
%\begin{figure}
%\begin{center}
%\includegraphics[width=\linewidth]{images/dog_models.png}
%\end{center}
%   \caption{The 8 different 3D dog models used to generate varied synthetic data.}
%\label{fig:dog_models}
%\end{figure}
%------------------------
\\\textbf{Dog textures.} To create different types of textures quickly and without the need of manually UV-unwrapping the models and manually producing textures, the textures were generated procedurally using shaders and mapped onto the surfaces by applying triplanar mapping. Triplanar mapping is a technique which applies textures onto a model from three directions using the world space positions. The initial setup of the shader can take time, but once implemented many textures can be generated by modifying the parameters of the shader such as the colour, size and position of the spots and the main colour of the dog. %Figure \ref{fig:texture_examples} illustrates different type of furs by varying the shader's parameters. 
In total we have generated 12 types of fur texture, which are randomly sampled when rendering the images.
% \begin{figure}
% \begin{center}
% \includegraphics[width=\linewidth]{images/different_textures.png}
% \end{center}
%   \caption{Examples of different type of fur textures procedurally generated using shaders. Implementing the shader can take time, but once it is done, different textures can be generated by modifying parameters such as the colour, size and position of the spots and the main colour of the dog.}
% \label{fig:texture_examples}
% \end{figure}
%------------------------
\\\textbf{Post-processing effects.} The post-processing effects from Unity were used to generate different lighting conditions and add noise to the renders. We added different types of grain which differ in particle size, intensity value, colour, and luminance contribution. We colour graded the image with saturation values which are randomly sampled between [-100, 100]; and with brightness values that range between [-20, 35]. \\
%------------------------
\textbf{Camera.} The camera was set to follow and look at the dog while being randomly positioned around the dog to capture it from different angles. The camera's field of view values were sampled uniformly at random between [50,100] degrees.\\
%------------------------
\textbf{Environment.} Different environments were created by modifying the sky and terrain textures. To set the sky texture we randomly sampled 1341 images from the Kaggle Landscape Pictures dataset \cite{landscapedataset}. %which contains landscapes, mountains, seas, deserts, islands and a category that they refer to as \textit{Japan}. 
Additionally, 10 different terrain textures were collected from the internet consisting of grass, autumn leaves (2x), dry mud (2x), cobble stone, pebbles, sand, snow and tiles. %All textures were sampled randomly when rendering the images.\\
%------------------------
\\\textbf{2D annotation.} To save the 2D annotations we located the 3D joint positions in world space and transform them into screen space. When the program runs, the 2D keypoints, the frame number and the bounding box coordinates with the 256x256 RGB image are saved. The bounding box coordinates were computed by adding 10 pixels to the minimum and maximum of the $x$- and $y$-coordinates. The average time to produce a synthetic is 33 milliseconds. By contrast, manual annotation of an image with these keypoints typically takes at least a minute. The data was generated on a MacBook Pro 2016 with a 2.9 GHz Quad-Core Intel Core i7 processor and a AMD Radeon Pro 460 4GB. 
%-------------------------------------------------------------------------
%%%%% EXPERIMENTS %%%%%
%-------------------------------------------------------------------------
\section{Experiments and Results}
We trained a 2-stacked hourglass network with 2 blocks (2HG), an 8-stacked hourglass network with 1 block (8HG) and a pre-trained Mask R-CNN model with a ResNet50 as a backbone. %from the TorchVision library. 
%The stacked hourglass network consists of successive steps of pooling and upsampling layers (hourglass module) which are stacked together to produce a final set of predictions. It has the capability of preserving spatial information at every scale and it has knowledge of the relationship of adjacent joints due the intermediate supervision between the hourglass modules. The Mask R-CNN can deduce the skeleton configuration of multiple people in an image by detecting each person and locating their joint locations simultaneously. This made the network successful because it eliminated the need for a human detector and pose estimator separately.
We refer the reader to the supplementary material and \cite{DBLP:journals/corr/HeGDG17, DBLP:journals/corr/NewellYD16} for further details on the training set up and the networks, respectively.

Six experiments were conducted: Firstly, the networks were trained with solely synthetic data. Secondly, the networks were trained purely on the StanfordExtra dataset. Thirdly, the networks which were trained only on synthetic data were fine-tuned with the StanfordExtra dataset using the same parameters as the first experiment. Then, we repeated the third experiment but with a smaller learning rate. Finally, the networks were trained on a mixed dataset which is a dataset that contains both the StanfordExtra and (either the whole or a fraction of) the SyDog dataset.
%-----------------------------
\subsection{Datasets}
\textbf{The StanfordExtra dataset} \cite{biggs2020left} is based on the Stanford Dogs dataset \cite{KhoslaYaoJayadevaprakashFeiFei_FGVC2011} and contains 12k real images which cover 120 different types of dogs. The 2D joint annotations were modified to reflect our synthetic data labels. Only the common joints were included in the annotations and the joints that differed between the StanfordExtra and the Synthetic Dog datasets, the keypoints' visibility were set to invisible. We used the StandfordExtra training-test split, which are publicly available \cite{biggs2020wldo}.
To train the networks on synthetic data, \textbf{the SyDog dataset} was divided by the different types of dog. 6 dogs were used for training, 1 for validation and 1 for testing. To train the networks with the mixed dataset, either the whole or a fraction of SyDog dataset was made available for training.
%------------------------------
\subsection{Evaluation metrics}
The networks were evaluated using the percentage of correct keypoints (PCK) and the mean per joint position error (MPJPE), which were both normalized with respect to the length of the bounding box diagonal. The PCK measures whether the predicted keypoints are within a threshold from the true keypoints. The threshold was set to 10\% of the bounding box diagonal. The MPJPE is the mean of the per joint position error \cite{Li20143DHP}. The evaluation metrics are calculated for visible keypoints only.
%------------------------------
\subsection{Results for SyDog test dataset} %The 2-, 8-stacked hourglass network (2HG, 8HG) and Mask R-CNN were trained with different learning rates (see supplementary table), the best results are shown in 
Table \ref{table:synthetic_data} shows the pose estimation results for the 2HG, 8H and Mask R-CNN. Some challenges do arise for the Mask R-CNN when it has to predict certain poses such as sitting and when it is presented with certain camera view points such as when the dog is far away. 
\begin{table}
\begin{center}
\begin{tabular}{|l|c|c|}
\hline
Network & PCK (\%) & MPJPE (\%) \\
\hline\hline
2HG & 77.76 & 6.51 \\
8HG & 77.57 & 6.56 \\
Mask R-CNN & 68.98 & 11.02 \\
\hline
\end{tabular}
\end{center}
\caption{Average PCK@0.1 and MPJPE from the 2HG, 8HG and the Mask R-CNN on the SyDog test dataset. \label{table:synthetic_data}}
\end{table}
%-------------------------------------------------------------------------
\subsection{Results for StanfordExtra test dataset}
The average PCK and MPJPE for all experiments are shown in Table \ref{table:compare_data}.
%------TABULAR----------------
\begin{table*}[t]
\begin{center}
\scalebox{1.0}{
\begin{tabular}{|l|l|l|c|c|}
\hline
Network & Dataset & Learning rate & PCK (\%) $\uparrow$ & MPJPE (\%) $\downarrow$\\
\hline\hline
\multirow{7}{*}{2HG} & Real & 0.001 & 68.61 & 15.84 \\\cline{2-5}
& Synthetic & 0.001 & 16.20 & 46.26 \\
& FT & 0.001 $\rightarrow$ 0.001 & 76.57 & 11.80 \\
& FT & 0.001 $\rightarrow$ 0.000001 & \textbf{77.19} & \textbf{11.32} \\
& Mixed@0.1 & 0.001 & 63.14 & 19.08 \\ 
& Mixed@0.5 & 0.001 & 68.43 & 15.50 \\ 
& Mixed@1.0 & 0.001 & 70.46 & 14.76 \\ 
\hline\hline
\multirow{7}{*}{8HG} & Real & 0.001 & 68.90 & 15.64 \\\cline{2-5}
% & Real & 0.0025 & 71.60 & 14.84 \\\cline{2-5}
& Synthetic & 0.001 & 17.34 & 45.08 \\ 
% & Synthetic & 0.0025 & 11.33 & 44.95 \\ 
& FT & 0.001 $\rightarrow$ 0.001 & 78.31 & 11.47 \\ 
& FT & 0.001 $\rightarrow$ 0.00001 & \textbf{78.65} & \textbf{11.19} \\ 
% & FT & 0.0025 $\rightarrow$ 0.0025 & 72.05 & 14.38 \\ 
% & FT & 0.0025 $\rightarrow$ 0.0001 & 72.06 & 13.70 \\
& Mixed@0.1 & 0.001 & 65.04 & 17.81 \\ 
& Mixed@0.5 & 0.001 & 71.76 & 15.19 \\ 
& Mixed@1.0 & 0.001 & 72.09 & 14.97 \\ 
\hline\hline
\multirow{7}{*}{Mask R-CNN} & Real & 0.00001 & 43.60 & 21.58 \\\cline{2-5}
& Synthetic & 0.001 & 13.22 & 37.49 \\ 
& FT & 0.00001 $\rightarrow$ 0.00001 & \textbf{50.77} & \textbf{20.03} \\ 
& FT & 0.00001 $\rightarrow$ 0.000001 & 46.58 & 21.17 \\ 
& Mixed@0.1 & 0.001 & 41.27 & 22.82 \\ 
& Mixed@0.5 & 0.001 & 47.71 & 21.64 \\ 
& Mixed@1.0 & 0.001 & 45.77 & 21.61 \\ 
\hline
\end{tabular}}
\end{center}
\caption{Results on the StanfordExtra test dataset. Results are shown from the 2-and 8-stacked hourglass (2HG, 8HG) and the Mask R-CNN trained solely on the StanfordExtra dataset (Real) and solely on the SyDog dataset (Synthetic) together with the fine-tuned (FT) models and the models trained with a mixed dataset (Mixed@fraction). The performance is evaluated using the percentage of correct keypoints (PCK) with a threshold set to 0.1 and the mean per joint per error (MPJPE) which are both w.r.t. the length of the ground truth bounding box diagonal. \label{table:compare_data}}
\end{table*}
%------TABULAR----------------
\\\textbf{Isolated training} %When testing the networks which used only synthetic data in the training phase, it is observed that the models perform poorly, which was expected, due to the domain gap.
The networks trained solely on synthetic data performed poorly on real data, which was expected due to the domain gap. Another possible reason could be that the SyDog dataset does not cover all breeds in the StanfordExtra dataset. The results from the networks trained only on real data were used as a baseline to evaluate the use of the SyDog dataset.
\\\textbf{Fine-tuning}
There's a significant increase in performance when fine-tuning the stacked hourglass networks with the same learning rate, and there is an even better performance when fine-tuning with a smaller learning rate. When fine-tuning the 2HG and the 8HG with a smaller learning rate the models' PCK performances are increased by 12.51\% and 14.15\%, respectively. The performance of the Mask R-CNN did not improve when fine-tuning with smaller learning rates, yet it performs better than the Mask R-CNN that was trained solely on real data. %We refer the reader to look at the supplementary section to view the table which contains the results of the Mask R-CNN when fine-tuned with smaller learning rates. 
\\\textbf{Training with mixed dataset} The best performance for the stacked hourglass networks is when the mixed dataset contains the full synthetic dataset, this is different for the Mask R-CNN; the Mask R-CNN performs best when the mixed dataset contains only half of the synthetic dataset, however using the full synthetic dataset in the mixed dataset produces better results than when the Mask R-CNN is trained solely with real data. 
%-----------------------------------

Our results clearly demonstrate the benefits of using the synthetic data generated by our system. %Figure \ref{fig:statistics_pck} demonstrates the increase in PCK and Figure \ref{fig:statistics_mpjpe} demonstrates the decrease in MPJPE. 
We show that using a mixed dataset when training gives a slight boost in the performance and that fine-tuning the networks results in a significant boost in the performance compared to the networks trained only on real data. 
% Figure \ref{fig:2hg_comparison_qualitative} compares qualitatively the results from the N-stacked hourglass networks trained purely on real data, fine-tuned with real data and trained with the mixed dataset which contains the full synthetic data (@1.0). 
The synthetic data generated by our system is not very photorealistic. However, it already improves the accuracy of the pose estimation models by 12.51\% in the case of the 2-stacked hourglass network. We expect that improvements in photorealism would result in further improvements in pose estimation. 
%-------------------------------------------------------------------------
%%%% CONCLUSION %%%%%%%
%-------------------------------------------------------------------------
\section{Conclusion}
To solve the lack of animal datasets, we introduce SyDog a synthetic dataset containing dogs with 2D pose and bounding box annotation which was generated using real-time rendering technology. The dataset was made varied by modifying the dog's appearance, pose, environment, lighting conditions (post-process effects) and camera view points. To evaluate the use of the SyDog dataset we conducted extensive experiments on the real dataset, StanfordExtra. We bridged the domain gap by fine-tuning the networks trained on synthetic data with real data and training the networks with a mixed dataset (synthetic+real). We demonstrated that using the SyDog dataset \textbf{increases the performance of pose estimation models} trained solely on real data and significantly \textbf{reduces} the need for \textbf{labour intensive labelling} which in turn \textbf{speeds up the process}. The models trained with a mixed dataset return a slight increase in performance and models that were fine-tuned with real data return a significant increase in performance. %The performance for the 2- and 8-stacked hourglass network and the Mask R-CNN when fine-tuned with the same learning rate were increased by 11.60\%, 13.66\% and 16.45\%, respectively. 
The data generated does not look very photorealistic; however we showed that it already notably improves the accuracy of the pose estimation models. Future work would involve improving the photorealism of the data for yet further improvements in pose. In this work we focused on 2D pose estimation and generated data with 2D annotations but with further work, the system could be extended to generate 3D annotations and modified to produce scenes that handle occlusions, multiple dogs and interactions.

{\small
\bibliographystyle{ieee_fullname}
\bibliography{egbib}
}

\end{document}

%%%%%%%%% TITLE
\title{Supplementary Material\\ SyDog: A Synthetic Dog Dataset for Improved 2D Pose Estimation}

\author{Moira Shooter \qquad Charles Malleson \qquad Adrian Hilton \\
University of Surrey\\
Stag Hill, University Campus, Guildford GU2 7XH\\
{\tt\small \{m.shooter,charles.malleson,a.hilton\}@surrey.ac.uk}
}

\maketitle
%%% DATA GENERATION %%%
\section{Data Generation}
\begin{figure*}
\begin{center}
   \includegraphics[width=\linewidth]{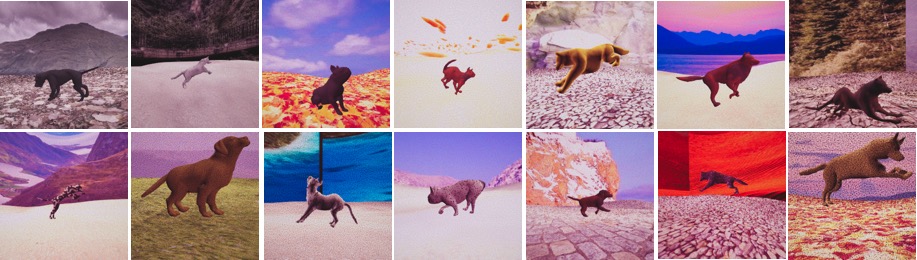}
\end{center}
   \caption{Examples of the SyDog dataset. The data is made varied using different lightning conditions (post-process effects), environments, dog's appearance and camera viewing points.}
\label{fig:synthetic_dataset_samples}
\end{figure*}

Figure \ref{fig:synthetic_dataset_samples} shows some example images from the SyDog dataset.

\section{Training Set Up}
We used a GeForce RTX 2080 Ti for training. The code was implemented with Pytorch Lightning \cite{falcon2019pytorch}. In this paper we focus on the usage of synthetic data and not the architecture design therefore we trained a 2-Stacked Hourglass network with 2 blocks (2HG), an 8-Stacked Hourglass network with 1 block (8HG) and a pre-trained Mask R-CNN model with a ResNet50 as a backbone from the TorchVision library. For 2HG and 8HG RMSprop was used as an optimiser with a learning rate set to \num{1e-3} and %additionally, the 8HG was trained with a learning rate set to \num{2.5e-3}; 
for the Mask R-CNN we also used an RMSprop but with a learning rate set to \num{1e-5}. The batch size was set to 32 for the stacked hourglass networks and 16 for the Mask R-CNN, although when the networks were trained with the mixed dataset, which consists of both real and synthetic samples, the batch size was set to 8, 4 of which were real samples and 4 of which were synthetic samples. We applied early stopping to our networks, the networks stop training when the validation does not improve for 10 epochs. The model is saved when there is an improvement in the validation loss. Originally the loss function for the stacked hourglass network would be the mean squared error between the ground truth $x_i$ and all the heatmaps generated by the network $y_i$ but because we only care about the visible keypoints, we modified the loss function by multiplying the keypoints' ground truth visibility $v_{i}$ with the squared error such that only the visible keypoints contribute to the loss function. 
\begin{equation}
    MSE_{masked} = \frac{1}{n}\sum_{i=1}^{n}v_{i}(y_{i} - x_{i})^{2}, v_{i}=0,1
\end{equation}
For the Mask R-CNN's loss function we sum the classification, regression and keypoint loss which our returned by the Mask R-CNN during training. 

In order to train the stacked hourglass network we represented the joints as 2D heatmaps. The ground truth heatmaps are produced by generating 2D Gaussians with a standard deviation (std) of 3 pixels centered at the joint's location. The stacked hourglass network takes 256x256 RGB-images as input and returns 25 heatmaps of size 64x64. Because the StanfordExtra dataset contains different sized images, it was necessary to resize them to 256x256; there was no need to resize the synthetic images as they were generated to be size 256x256. 
To train the Mask R-CNN the joints were represented as a list containing the joint's coordinates and visibility. The Mask R-CNN takes 256x256 images as input and returns the predicted bounding boxes, labels, scores of each prediction and the locations of the predicted keypoints.
Before feeding the data to the networks, we made sure that each pixel had the same similar data distrubtion by normalizing the data. The StanfordExtra dataset was normalised with a mean=[0.4822, 0.4621, 0.3972] and a std=[0.2220, 0.2172, 0.2167]; while the SyntheticDog dataset was normalised with a mean=[0.6528, 0.4980, 0.5418] and a std=[0.1827, 0.1970, 0.1946].
%--- TABLE ---- 
\begin{table}
\begin{center}
\begin{tabular}{|l|c|c|}
\hline
Learning rate & PCK $\uparrow$ (\%) & MPJPE $\downarrow$ (\%) \\
\hline\hline
    \num{1e-5} & \textbf{50.77} & \textbf{20.03} \\
    \num{1e-6} & 46.58 & 21.17 \\
    \num{1e-7} & 44.87 & 22.53 \\
    \num{1e-8} & 41.32 & 22.53 \\
    \num{1e-9} & 39.68 & 23.74 \\
\hline
\end{tabular}
\end{center}
\caption{Pose estimation results from the Mask R-CNN on the StanfordExtra test dataset when fine-tuned with a smaller learning rate. The performance is evaluated using the percentage of correct keypoints (PCK) with a threshold set to 0.1 and the mean per joint per error (MPJPE) which are both normalised w.r.t. the length of the ground truth bounding box diagonal.\label{table:mask_rcnn_fine_tuned}}
\end{table}
%--- TABLE ---- 
%%%% EXPERIMENTS AND RESULTS %%%%
\section{Experiments and Results}
Table \ref{table:mask_rcnn_fine_tuned} shows the pose estimation results from the Mask R-CNN when fine-tuned with smaller learning rates. 
% \section{Qualitative Results}
\begin{figure}
\subfloat[Bar graph of the percentage of correct keypoints (PCK) between various pose estimation models and training data.]{
\begin{minipage}{
	   0.45\textwidth}
	     \centering
	     \includegraphics[width=0.85\linewidth]{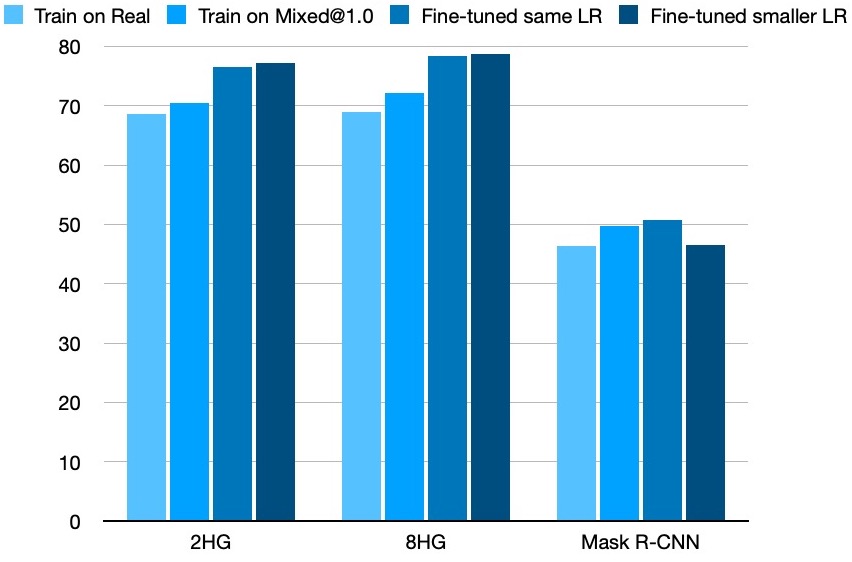}
	     \label{fig:statistics_pck}
\end{minipage}
}
\hfill
\subfloat[Bar graph of the mean per joint per error (MPJPE) between various pose estimation models and training data.]{
\begin{minipage}{
	   0.45\textwidth}
	     \centering
	     \includegraphics[width=0.85\linewidth]{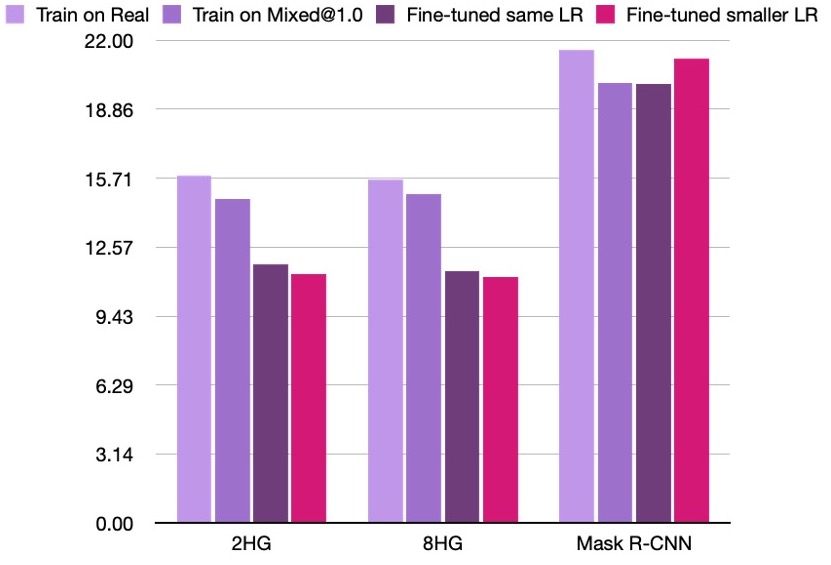}
	     \label{fig:statistics_mpjpe}
\end{minipage}
}
  \caption{Quantitative comparison on StanfordExtra test dataset for the 2HG, 8HG and Mask R-CNN trained purely on real data, fine-tuned with real data and trained with the mixed dataset.}
\end{figure}

\begin{figure*}
\begin{center}
  \includegraphics[width=\linewidth]{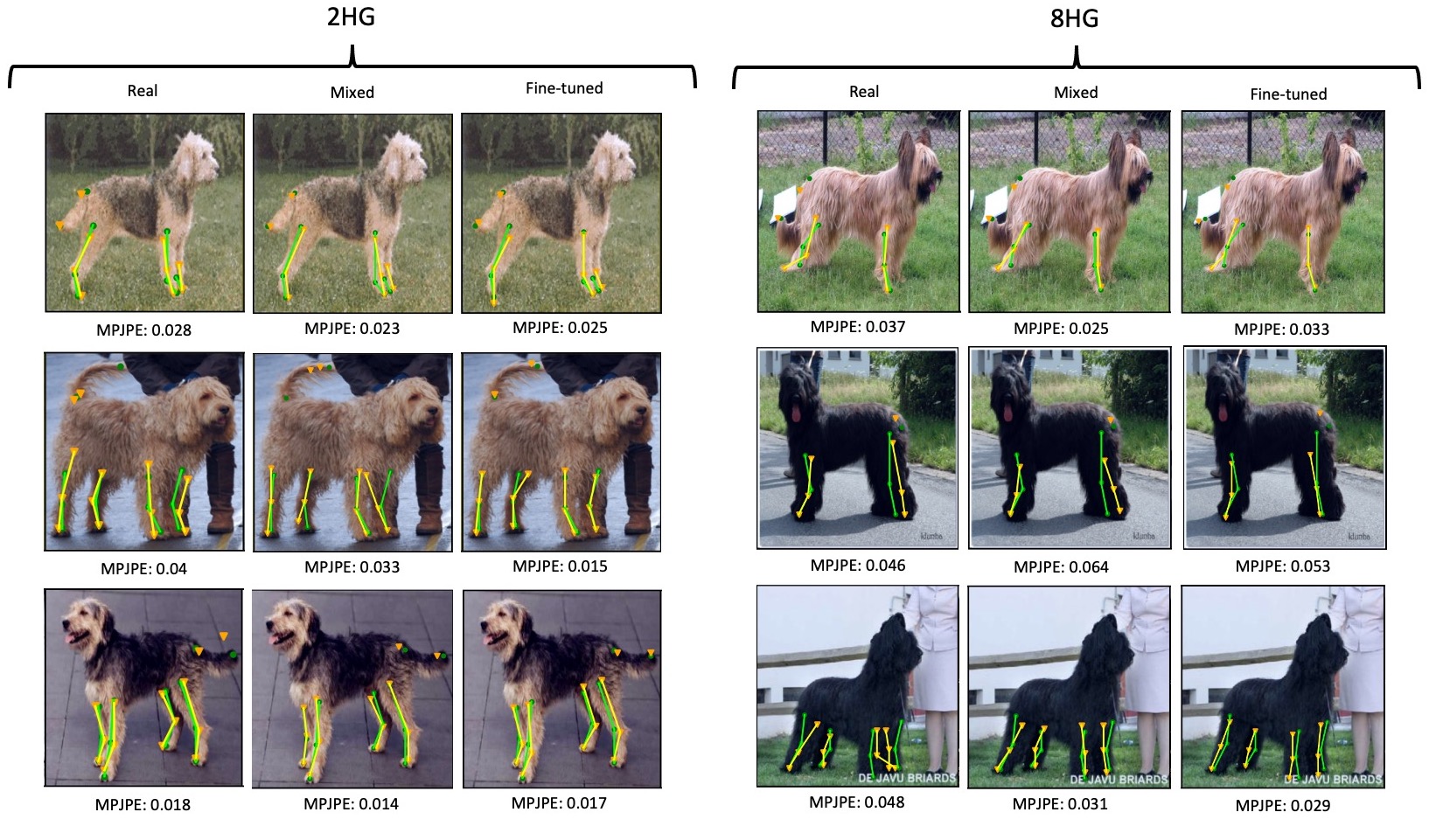}
\end{center}
  \caption{Qualitative comparison on StanfordExtra between the N-stacked hourglass networks trained purely on real data, fine-tuned with real data and trained with the mixed dataset. The ground truth (green) and predicted (yellow) pose with the mean per joint per error (MPJPE) are displayed.}
\label{fig:2hg_comparison_qualitative}
\end{figure*}

{\small
\bibliographystyle{ieee_fullname}
\bibliography{egbib}
}